\documentclass{article}
\usepackage{styles/spconf}
\usepackage{styles/defs}

\usepackage{amsmath,graphicx}
\usepackage{xcolor}
\hypersetup{hidelinks}

\title{Physics-Informed Sylvester Normalizing Flows for Bayesian Inference in Magnetic Resonance
Spectroscopy}

\name{%
  Julian P. Merkofer$^{\star, \dagger}$%
  \qquad Dennis M. J. van de Sande$^{\star}$%
  \qquad Alex A. Bhogal$^{\dagger}$
  \qquad Ruud J. G. van Sloun$^{\star}$
  \thanks{This work was in part funded by Spectralligence (EUREKA IA Call, ITEA4 project 20209) and NWO VIDI (VI.Vidi.223.085). We also acknowledge support from the NVIDIA Academic Hardware Grant Program.}%
}
\address{%
  $^{\star}$ Eindhoven University of Technology, Eindhoven, The Netherlands \\%
  $^{\dagger}$ University Medical Center Utrecht, Utrecht, The Netherlands%
}

\begin{document}
\ninept

\maketitle

%%%%%%%%%%%%%%%%
%%% Abstract %%%
%%%%%%%%%%%%%%%%
\begin{abstract}
% 100-150 words
\Ac{mrs} is a non-invasive technique to measure the metabolic composition of tissues, offering valuable insights into neurological disorders, tumor detection, and other metabolic dysfunctions. However, accurate metabolite quantification is hindered by challenges such as spectral overlap, low signal-to-noise ratio, and various artifacts. Traditional methods like linear-combination modeling are susceptible to ambiguities and commonly only provide a theoretical lower bound on estimation accuracy in the form of the \acl{crb}. This work introduces a Bayesian inference framework using \acp{snf} to approximate posterior distributions over metabolite concentrations, enhancing quantification reliability. A physics-based decoder incorporates prior knowledge of \ac{mrs} signal formation, ensuring that the learned posteriors are physically meaningful. We validate the method on simulated 7T proton \ac{mrs} data, demonstrating accurate metabolite quantification, well-calibrated uncertainties, and insights into parameter correlations and ambiguities.

\end{abstract}
\begin{keywords}
Normalizing Flows, Bayesian Inference, Physics-Informed Deep Learning, Magnetic Resonance Spectroscopy
\end{keywords}

%\newcommand{\cem}[1]{\textcolor{blue}{cem: #1}}

%%%%%%%%%%%%%%%%%%%%
%%% Introduction %%%
%%%%%%%%%%%%%%%%%%%%
\vspace{-1mm}
\section{Introduction} \label{sec:intro}
\vspace{-2mm}

% reset abbrevs after abstract
\acresetall

Nuclear \ac{mrs} is a powerful, non-invasive technique that extends the diagnostic capabilities of conventional \ac{mri} by providing quantitative biochemical information about tissue composition in vivo \cite{Condon2011MagneticRI}. While traditional \ac{mri} primarily reveals structural and anatomical features, \ac{mrs} enables direct measurement of metabolite concentrations, offering insights into underlying cellular metabolic processes \cite{Maudsley2020AdvancedMR}. This metabolic profiling is especially valuable for the early diagnosis, monitoring, and treatment of neurological disorders, cancer, and other metabolic dysfunctions \cite{faghihi_magnetic_2017, Horska2023MRSClincalA}.

Despite its significant potential, the broader clinical adoption of \ac{mrs} is limited by persistent challenges in metabolite quantification. The standard approach, known as linear-combination modeling, fits the measured spectrum as a weighted sum of basis functions representing the known spectral signatures of metabolites, typically using least-squares optimization~\cite{near_preprocessing_2021}. However, the method suffers from inherent ambiguities: multiple metabolite combinations can produce similarly plausible fits, particularly given the spectral overlap of metabolites in proton \ac{mrs}, where different metabolites contribute signals at similar resonance frequencies. The accuracy of estimates is further compromised by inherently low \ac{snr}, baseline distortions, and numerous artifacts \cite{Kreis2004IssuesOS, Hurd2009ArtifactsAPI}. Although uncertainty is often characterized using the \ac{crb}, it merely provides a theoretical lower bound on the variance of unbiased estimators and does not capture parameter interdependencies or model mismatch \cite{Landheer2021AreCRLBs}.

Bayesian inference offers a more comprehensive framework: instead of point estimates, it yields a posterior distribution over the parameters of interest, such as metabolite concentrations and other spectral characteristics, providing a probabilistic representation of the information contained in the acquired spectrum. %This enables a more nuanced understanding of the metabolite signals, accounting for complex dependencies and trade-offs between metabolites. Moreover, it can capture a potential multi-modal nature of specific parameters, revealing ambiguities in the fitting process that may be crucial for clinical decision-making.
Since the exact posterior is generally intractable, we approximate it with normalizing flows, which transform a simple base distribution through a sequence of invertible transformations with tractable Jacobians \cite{Kobyzev2020NormalizingFA}. Specifically, we employ \acp{snf} \cite{vandenBerg2018SNFs}, which generalize planar flows by overcoming their single-unit bottleneck. %\Acp{snf} leverage Sylvester's determinant identity to ensure that the Jacobian determinant of the transformation can be computed efficiently \cite{vandenBerg2018SNFs}.

% \begin{figure*}
%     \centering
%     \vspace{-1mm}
%     \includegraphics[width=1.95\columnwidth]{figures/spectra.pdf}
%     \vspace{-3mm}
%     \caption{
%     %Illustration of key challenges in \ac{mrs} metabolite quantification. From left to right, the figure shows simulated proton \ac{mrs} spectra: a noiseless spectrum decomposed into individual metabolite components (colored), highlighting substantial spectral overlap due to limited spectral resolution; the same spectrum with broadened peaks (also noiseless), simulating poor B0 shimming that exacerbates overlap and further obscures metabolite features; and a spectrum with reduced \ac{snr}, demonstrating how noise can mask low-intensity metabolites and complicate accurate quantification (\ac{snr} = 7 dB).
%     Illustration of key challenges in \ac{mrs} metabolite quantification. From left to right: a noiseless spectrum with individual metabolite components, showing substantial spectral overlap; the same spectrum with broadened peaks due to poor B0 shimming, increasing overlap; and a low-\ac{snr} spectrum (\ac{snr} = 7 dB), where noise obscures low-intensity metabolites.
%     }
%     \label{fig:spectra}
% \end{figure*}

In this work, we apply \acp{snf} to Bayesian inference in \ac{mrs}, aiming to improve metabolite quantification by constructing flexible posterior distributions that reveal parameter interdependencies and provide calibrated uncertainty estimates. Our model integrates a physics-based decoder, ensuring that the learned representations correspond to metabolite concentrations and other physical parameters of the \ac{mrs} signal, and allowing prior knowledge about the signal and the spectral properties of metabolites to be incorporated. We train and validate our approach using simulated proton \ac{mrs} data of the human brain, where ground truth is available for comparison. Results demonstrate that our method yields accurate metabolite quantification with well-calibrated uncertainties, while revealing interdependencies and ambiguities between parameters. %These findings offer a deeper understanding of \ac{mrs} data, advancing the reliability and interpretability of metabolite quantification in clinical settings.

%%%%%%%%%%%%%%%
%%% Methods %%%
%%%%%%%%%%%%%%%
\vspace{-1mm}
\section{Methods} \label{sec:methods}
\vspace{-2mm}

\begin{figure*}
    \centering
    \vspace{-1mm}
    \includegraphics[width=2\columnwidth]{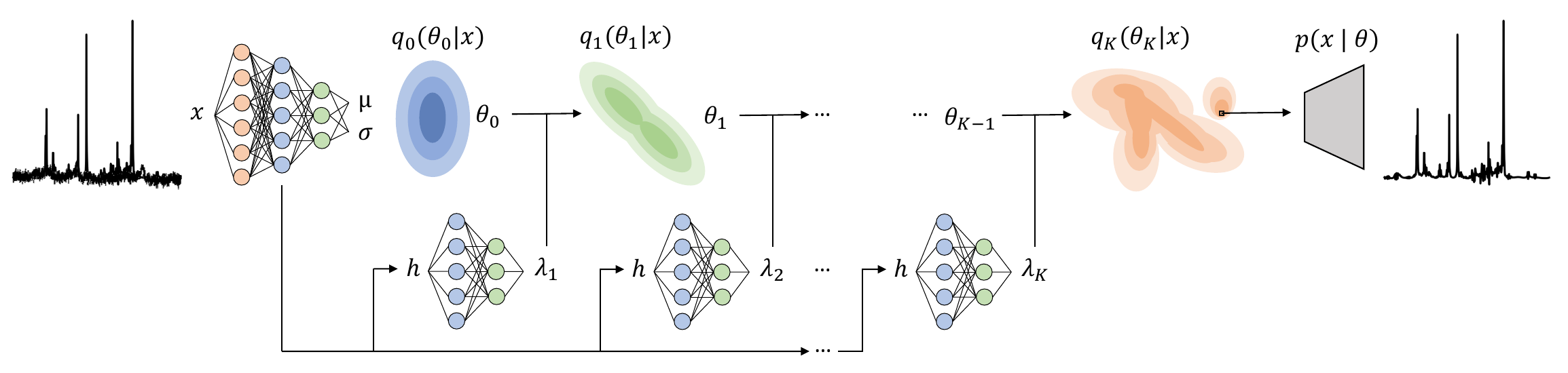}
    \vspace{-3mm}
    \caption{The inference network takes the observed spectrum $x$ as input and outputs the parameters of the base distribution $q_0(\theta_0 \,|\,x)$ for the latent variables $\theta$. \Acp{snf} transform the base distribution into a more complex and flexible posterior $q_K(\theta_K \,|\,x)$. %Instead of a decoder network, the physics-based \ac{mrs} signal model directly maps the latent variables $\theta_K$, representing metabolite concentrations and other relevant parameters, to the spectral domain.
    Instead of a decoder network, the likelihood $p(x \,|\, \theta)$ is given by the physics-based \ac{mrs} signal model of Eq.~\eqref{eq:signal_model}, which directly maps the latent variables $\theta_K$, representing metabolite concentrations and other relevant parameters, to the spectral domain; this block is a fixed model, not a learned network.
    \vspace{-2mm}}
    \label{fig:model}
\end{figure*}

\subsection{Problem Formulation}
\vspace{-1mm}
The central challenge in \ac{mrs} metabolite quantification lies in the ill-posed nature of the inverse problem, arising from factors such as inherent spectral overlap due to limited resolution, signal degradation from noise and artifacts, and the presence of unmodeled baseline distortions. 
%To illustrate these challenges, Fig. \ref{fig:spectra} presents simulated proton \ac{mrs} spectra, highlighting key issues. 
Our goal is to move beyond the point estimates provided by traditional methods like LCModel \cite{provencher_estimation_1993} towards a full Bayesian treatment, i.e., to learn the conditional distribution $p(\theta \,|\,x)$, where $\theta$ is the vector of metabolite concentrations and other signal parameters, and $x$ is the observed \ac{mrs} signal. %This allows us to estimate not only the most likely concentrations but also the inherent variability within these estimates, which is crucial for reliable interpretation.

\vspace{-1mm}
\subsection{Physics-Based Forward Model}
\vspace{-1mm}
We leverage a physics-based forward model $p(x \,|\, \theta)$, which describes the generation of an \ac{mrs} signal $x$ from a set of underlying parameters $\theta$. The parameter vector is defined as
\begin{equation}
    \theta = \{a_1, \ldots, a_M, \gamma, \varsigma, \epsilon, \phi_0, \phi_1, b_1, \ldots, b_{2(L-1)}\},
\end{equation}
where $a_m$ are the metabolite concentrations, $\gamma$ and $\varsigma$ are linewidth parameters, $\epsilon$ is a frequency shift, $\phi_0$ and $\phi_1$ are phase terms, and $b_1, \ldots, b_{2(L-1)}$ define a complex $L$-order polynomial baseline.

Although $x$ is acquired in the time domain, we express the forward model in the frequency domain, where spectral features are captured more conveniently. The expected spectrum given $\theta$ is
\begin{equation} \label{eq:signal_model}
X(f \,|\, \theta) = e^{i(\phi_0 + f\phi_1)} \sum_{m=1}^{M} a_m \mathcal{F}\left\{s_m(t) e^{-(i\epsilon + \gamma + \varsigma t)t}\right\} + B(f),
\end{equation}
where $\mathcal{F}$ denotes the Fourier transform, $\{s_m(t)\}_{m=1}^M$ are the time-domain basis functions representing the spectral contributions of each metabolite (typically obtained using density matrix simulations \cite{Zhang2017FastSpectroscopy}), and $B(f)$ models the baseline.

\vspace{-1mm}
\subsection{Variational Inference with SNFs}
\vspace{-1mm}
To approximate the intractable true posterior distribution $p(\theta \,|\,x)$, we use a \ac{snf} model to learn an approximate posterior $q(\theta \,|\,x)$. The model architecture, outlined in Fig. \ref{fig:model}, consists of a probabilistic encoder, as in a \ac{vae} \cite{Kingma2014VAE}, combined with Householder Sylvester flows \cite{vandenBerg2018SNFs}; all components are implemented as fully connected networks.

The encoder network $q_{\psi}$ takes the observed spectrum $x$ as input and outputs the parameters of a base latent Gaussian, a mean vector $\mu(x) \in \mathbb{R}^{D_z}$ and a standard deviation vector $\sigma(x) \in \mathbb{R}^{D_z}$, as well as the flow parameters $\lambda$ \cite{Rezende2015normFlows}.
A latent sample $\theta_0$ is first drawn from the base Gaussian using the reparameterization trick:
\begin{equation}
    \theta_0 = \mu(x) + \sigma(x) \odot \eta, \quad \eta \sim \mathcal{N}(0, I),
\end{equation}
where $\odot$ denotes element-wise multiplication. To increase the flexibility of the approximate posterior, a sequence of $K$ \ac{snf} transformations is applied. Each transformation $f_k$ is parameterized by $\lambda_k(x)$ and transforms $\theta_{k-1}$ into $\theta_k$, producing a final latent variable $\theta_K \in \mathbb{R}^{D_z}$.

% Optional (if space allows): Sylvester flows suit this problem well: the latent dimension is small, all flow parameters are amortized by the encoder, and the log-determinant is computed in closed form independent of the spectrum length.
Sylvester flows use structured weight matrices and Sylvester’s determinant identity to compute the Jacobian determinants efficiently, yielding the approximate posterior
\begin{equation}
q_K(\theta_K \,|\, x) = q_0(\theta_0 \,|\, x) \prod_{k=1}^K \left| \det\left( \frac{\partial \theta_k}{\partial \theta_{k-1}} \right) \right|^{-1}.
\end{equation}

The model is trained by maximizing the \ac{elbo}. For an observation $\hat{x}$, the \ac{elbo} is:
\begin{equation}
\begin{split}
    &\mathcal{L}(\psi; \hat{x}) = \mathbb{E}_{ q_\psi(\theta_{0:K}|\hat{x})}
    \Big[
    \log \underbrace{p(x=\hat{x} \,|\, \theta_K)}_{\text{Physics Forward Model}} \\ & \underbrace{- \log  \underbrace{q_\psi(\theta_0 \,|\, \hat{x})}_{\text{Likelihood } \textit{Enc}_\psi}
    - \log \underbrace{p(\theta_K)}_{\text{Likelihood }\theta_K}
    + \underbrace{\sum_{k=1}^K \log \Big| \det\left(J\right)\Big|}_{\text{LogDet Jacobian Flows}}}_{\text{KL Divergence Posterior - Prior}}
    \Big],
\end{split}
\label{eq:elbo}
\end{equation}
where $J = J(\theta_k, \lambda_k(\hat{x}))$ denotes the Jacobian of flow layer $k$. %Before applying the physics-based forward model $p(x \,|\, \theta)$, the final latent variable is mapped through an affine transformation accounting for the prior information on the metabolites and signal parameters.
Note that, while the training data are simulated from uniform distributions over realistic parameter ranges (Sec.~\ref{sec:results}), the prior $p(\theta_K)$ in Eq.~\eqref{eq:elbo} is a Gaussian distribution with diagonal covariance whose standard deviations reflect these ranges; this choice was found to be essential for stable training and remains applicable when training on in vivo data, where the true parameter distribution is unknown.
At inference, $S$ samples $\theta_0$ are drawn from the base Gaussian conditioned on the observed spectrum and each is passed through the flow layers, yielding $S$ samples from the approximate posterior $q_K(\theta_K \,|\, \hat{x})$, from which the concentration estimate (posterior mean), credible intervals, and parameter correlations are obtained.

% \begin{equation}
%     \theta' = \mu_p + \sigma_p \odot \theta_K.
% \end{equation}

%%%%%%%%%%%%%%%
%%% Results %%%
%%%%%%%%%%%%%%%
\vspace{-1mm}
\section{Results} \label{sec:results}
\vspace{-2mm}

% Figure sources placed here (encountered on page 2) so that Fig. 2 becomes a float page (p. 3) and Fig. 3 sits at the top of p. 4.
% Composite figure from Test/t1k_e1k_s1k/SNFVAE_b10/aumc2/, example index 4 (Cr-PCr corr -0.79, GPC-PCh -0.76).
\begin{figure*}
    \centering
    \vspace{-1mm}
    \includegraphics[width=0.95\textwidth]{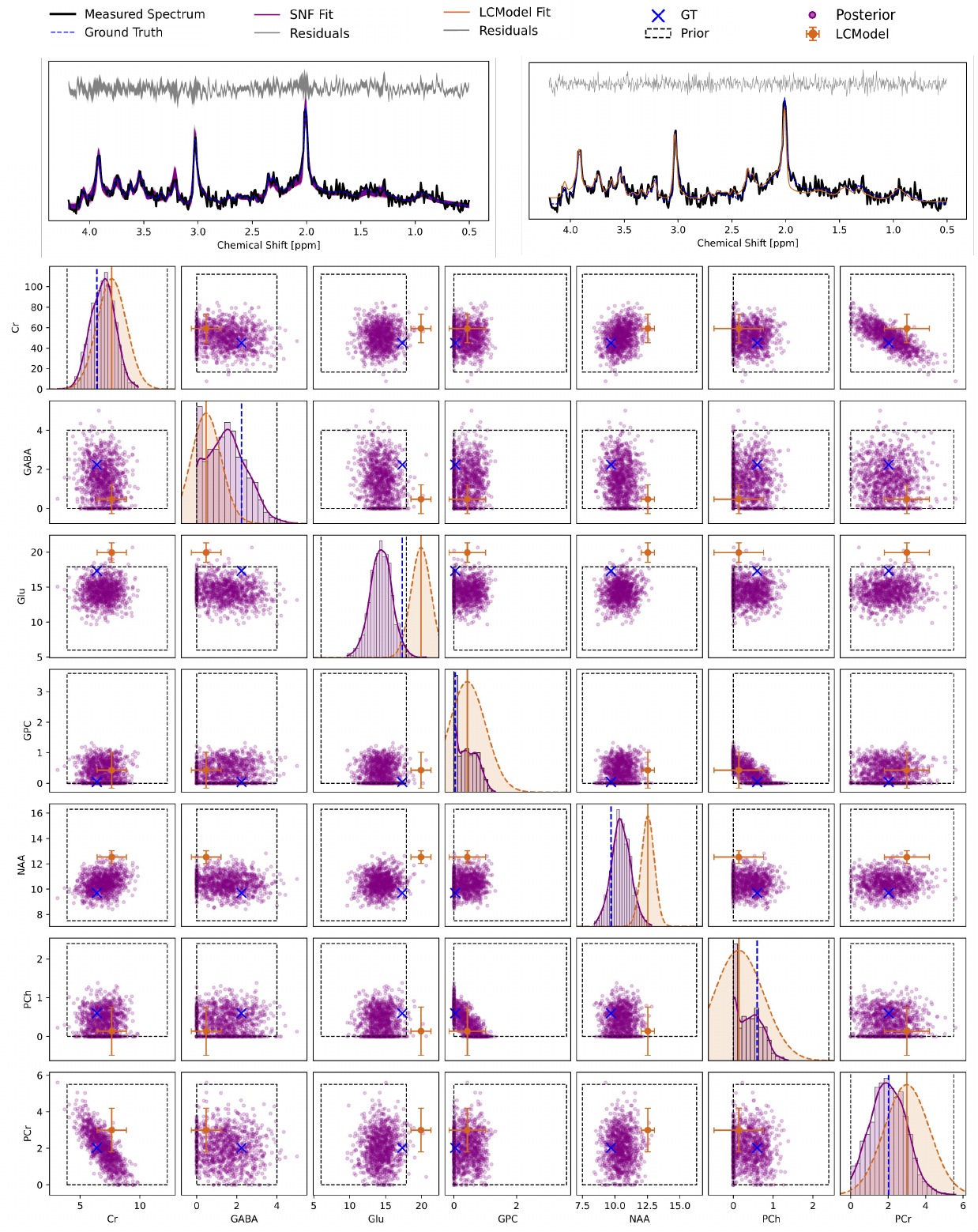}
    \vspace{-3mm}
    \caption{
    %Visualization of a test example spectrum $\hat{x}$, alongside its fits and latent representations. Top left: the reconstructed spectrum obtained by sampling 1000 latent variables $\theta_0 \sim \mathcal{N}(\mu(\hat{x}), \sigma(\hat{x}))$, transforming through the flows, and decoding to the corresponding spectra. Top right: LCModel fit of the same spectrum, shown for comparison. Bottom: a pairplot of the latent posterior distributions inferred by the \ac{snf}, showing individual marginal distributions along the diagonal and correlations between parameters off-diagonal. Dashed boxes indicate the prior distributions for each parameter. LCModel estimates and \acp{crb} are also shown for reference.
    Example test spectrum $\hat{x}$ fitted with the \ac{snf} and LCModel. Top left: reconstructed spectrum obtained by sampling from the \ac{snf} posterior. Top right: LCModel fit for the same spectrum. Bottom: pairplot of \ac{snf}-inferred posterior distributions for \ac{cr}, \ac{gaba}, \ac{glu}, \ac{gpc}, \ac{naa}, \ac{pch}, and \ac{pcr}. Marginal distributions are shown along the diagonal; correlations are shown off-diagonal. The uniform simulation ranges of the concentrations are indicated by dashed boxes; ground truth (blue) as well as LCModel estimates and \acp{crb} (orange) are overlaid for reference.
    }
    \label{fig:fits_n_latent}
\end{figure*}

\begin{figure*}
    \centering
    \vspace{-1mm}
    \includegraphics[width=0.95\linewidth]{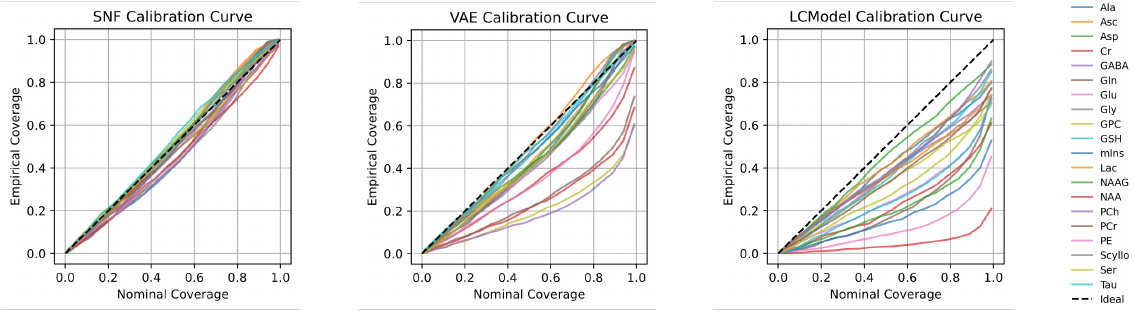}
    \vspace{-3mm}
    \caption{Calibration curves for \ac{snf} (left), \ac{vae} (middle), and LCModel (right), depicting nominal versus empirical coverage for each of the 20 metabolites across 1000 test spectra. The ideal calibration is shown as a dashed line, with nominal coverage ranging from 1\% to 99\%.\vspace{-2mm}}
    \label{fig:calibration}
\end{figure*}

\subsection{Experimental Setup}
\vspace{-1mm}
% metion model details
%% layers
%% flows
%% training
The architecture of the \ac{snf} model\footnote{The source code can be found online at \url{https://github.com/julianmer/SNFs-for-Bayesian-Inference-in-MRS}.} is based on a fully connected \ac{vae} and a sequence of Householder Sylvester flows. The encoder consists of three fully connected layers with 512, 256, and 128 units, followed by $K = 8$ Householder Sylvester flow layers with a hidden width of 128. The \ac{kl} term of the \ac{elbo} (Eq.~\eqref{eq:elbo}) is weighted by $\beta = 10$ to balance reconstruction accuracy and latent regularization \cite{Higgins2017betaVAE}.

% mention data details
%% sLaser, other seq params
%% number of metabs

The training and validation data consist of simulated 7T proton \ac{mrs} spectra of the human brain, generated with the signal model of Eq.~\eqref{eq:signal_model} using a semi-LASER basis set (TE = 34 ms, 1024 points, 3000 Hz bandwidth) of 20 metabolites and a macromolecular baseline. Metabolite concentrations are sampled uniformly from ranges spanning the union of literature values \cite{de_graaf_vivo_2019} and LCModel and FSL-MRS fits of in vivo 7T data; linewidth, frequency shift, phase, and baseline parameters are likewise sampled from uniform ranges covering realistic in vivo variability. Additive noise yields \acp{snr} between 5 and 63 dB. All models are evaluated on 1000 independently simulated test spectra from the same distribution, using $S = 1000$ posterior samples per spectrum.

Training is conducted by dynamically generating batches of 16 spectra per iteration, rather than relying on a fixed pre-generated dataset. Every 256 training batches, a new validation set of 1024 spectra is simulated, ensuring a roughly 20/80\% validation-to-training split. Optimization uses the Adam optimizer with a learning rate of $1\times10^{-4}$. Since data are simulated ad-hoc, the validation loss decreases smoothly and training proceeds until it converges; the checkpoint with the lowest validation loss is selected as the final model. For reconstruction, we employ \ac{mse} loss between the predicted spectra and the true spectra in the frequency range of interest (0.5 to 4.2 ppm).

% We compare three methods:
% \begin{itemize}
% \item \textbf{LCModel}: a widely used classical fitting approach in \ac{mrs} based on linear combination modeling.
% \item \textbf{VAE}: our model without normalizing flows, trained identically to the SNF.
% \item \textbf{SNF}: the full model with Householder Sylvester flows.
% \end{itemize}
We compare three methods: LCModel \cite{provencher_estimation_1993}, a widely used linear-combination modeling approach; VAE, our model without normalizing flows, trained identically; and SNF, the full model with Householder Sylvester flows.

\vspace{-1mm}
\subsection{Simulation Results}
\vspace{-1mm}

% % Numbers from Test/t1k_e1k_s1k/{SNFVAE_b10,baseVAE_b10}/aumc2/metrics/ (1000 test spectra, 1000 posterior samples, beta=10).
% \begin{table*}
% \centering
% \caption{Performance comparison across models. Metrics include \ac{mae}, negative \ac{elbo}, \ac{rss}, \ac{kl}, and \ac{ccc}.}
% \vspace{2mm}
% \label{tab:model_performance}
% \resizebox{2\columnwidth}{!}{%
% \begin{tabular}{lccccc}
% \toprule
% \textbf{Model} & \textbf{MAE} ↓ & \textbf{-ELBO} ↓ & \textbf{RSS} ↓ & \textbf{KL} ↓ & \textbf{CCC} ↑ \\
% \midrule
% SNF       &  \textbf{0.605 (± 0.018)}  &  \textbf{1567.010 (± 34.962)}  &  \textbf{1366.109 (± 34.786)}  &  \textbf{20.090 (± 0.139)}  &  \textbf{0.662 (± 0.008)}  \\
% % IAF       &  --  &  --  &  --  &  --  &  --  \\
% % VAE evaluated on the SAME 1000 test spectra as SNF/LCModel (Test/t1k_e1k_s1k_same/baseVAE_b10/aumc2/metrics/metrics.txt)
% VAE       &  0.608 (± 0.018)  &  1630.893 (± 35.237)  &  1395.813 (± 35.042)  &  23.508 (± 0.150)  &  0.660 (± 0.008)  \\
% Prior     &  2.354 (± 0.069)  &  1.38e14 (± 8.80e13)  &  1.38e14 (± 8.80e13)  &  8.61e6 (± 1.61e5)  &  2.91e-4 (± 6.52e-5)  \\
% LCModel   &  1.135 (± 0.039)  &  --  &  --  &  --  &  0.659 (± 0.005)  \\
% \bottomrule
% \end{tabular}
% }
% \vspace{-2mm}
% \end{table*}

% NOTE: the metrics files report the beta-weighted training loss (RSS + 10*KL) as 'Total Loss' (1567.010 / 1630.893).
% The table shows the proper -ELBO = RSS + KL; its SE uses Cov(RSS,KL) recovered from the three reported SEs.
\begin{table}[b]
\centering
\vspace{-3mm}
\caption{Performance comparison across models: \ac{mae} of the posterior-mean concentrations over the metabolites, negative \ac{elbo} (= \ac{rss} + \ac{kl}), spectral reconstruction error (\ac{rss}), and \ac{kl} divergence to the prior; mean $\pm$ standard error over 1000 test spectra.}
\vspace{1mm}
\label{tab:model_performance}
\resizebox{\columnwidth}{!}{%
\begin{tabular}{lcccc}
\toprule
\textbf{Model} & \textbf{MAE} ↓ & \textbf{-ELBO} ↓ & \textbf{RSS} ↓ & \textbf{KL} ↓ \\
\midrule
SNF       &  \textbf{0.605 (± 0.018)}  &  \textbf{1386.199 (± 34.801)}  &  \textbf{1366.109 (± 34.786)}  &  \textbf{20.090 (± 0.139)}  \\
% IAF       &  --  &  --  &  --  &  --  \\
% VAE evaluated on the SAME 1000 test spectra as SNF/LCModel (Test/t1k_e1k_s1k_same/baseVAE_b10/aumc2/metrics/metrics.txt)
VAE       &  0.608 (± 0.018)  &  1419.321 (± 35.059)  &  1395.813 (± 35.042)  &  23.508 (± 0.150)  \\
Prior     &  2.354 (± 0.069)  &  1.38e14 (± 8.80e13)  &  1.38e14 (± 8.80e13)  &  8.61e6 (± 1.61e5)  \\
LCModel   &  1.135 (± 0.039)  &  --  &  --  &  --  \\
\bottomrule
\end{tabular}
}
\vspace{-2mm}
\end{table}

% Shortened for the 4-page limit; original:
% To qualitatively assess the behavior of the inferred latent posteriors, we visualize in Fig.~\ref{fig:fits_n_latent} an example test spectrum and its corresponding fits. The \ac{snf} model samples 1000 latent variables $\theta_0 \sim \mathcal{N}(\mu(\hat{x}), \sigma(\hat{x}))$, which are then transformed through the flow layers to produce $\theta_K$, from which reconstructed spectra $p(x ,|, \theta)$ are generated. The top left panel shows the reconstructed spectrum from the \ac{snf}, while the top right panel displays the LCModel fit for the same measurement. Below, we visualize the posterior distributions inferred by the \ac{snf} for key metabolites, including:  \ac{cr}, \ac{gaba}, \ac{glu}, \ac{gpc}, \ac{naa}, \ac{pch}, and \ac{pcr}. The diagonal of the pairplot shows the marginal distributions, while the off-diagonal elements show pairwise correlations. Prior distributions are indicated by dashed boxes, and LCModel estimates along with \acp{crb} are overlaid for comparison.
To qualitatively assess the inferred posteriors, Fig.~\ref{fig:fits_n_latent} shows an example test spectrum with its LCModel and \ac{snf} fits (1000 posterior samples $\theta_K$ decoded through the forward model), and the posterior over \ac{cr}, \ac{gaba}, \ac{glu}, \ac{gpc}, \ac{naa}, \ac{pch}, and \ac{pcr}.
We observe that metabolites such as \ac{naa} exhibit tight, well-constrained distributions, indicating confident and unambiguous quantification. In contrast, broader distributions are seen for \ac{gaba}, reflecting higher uncertainty due to its inherently low concentration. Most notably, the strong anti-correlations between \ac{cr} and \ac{pcr} as well as between \ac{gpc} and \ac{pch} (correlation coefficients of $-0.79$ and $-0.76$ in this example) highlight the challenge of disentangling these highly overlapping signals: the spectrum constrains their sums well, but not their individual contributions, as visible in the elongated diagonal shape of their joint posteriors. While the \ac{crb} provides a theoretical lower bound on the variance of an estimator, it does not capture this structure; for \ac{cr}, \ac{pch}, and \ac{pcr} the LCModel \acp{crb} in this example span essentially the full simulation range. This structure is systematic: across the 1000 test spectra, the median \ac{snf} posterior correlation is $-0.79$ for \ac{cr}--\ac{pcr} and $-0.85$ for \ac{gpc}--\ac{pch} (below $-0.5$ in every spectrum), whereas it is zero for the \ac{vae} by construction. As a result, the \ac{snf} reports the sum \ac{cr}\,+\,\ac{pcr} with a median relative uncertainty of 6\% but the individual metabolites with 12\% and 36\%, while the \ac{vae} assigns them 5\% and 16\%, an over-confidence confirmed by the calibration analysis below.

Table~\ref{tab:model_performance} summarizes the quantitative performance across models. The \ac{snf} achieves the best scores on all metrics, indicating a tighter fit to the data at a smaller divergence from the prior than the \ac{vae}. While the \ac{vae} is close in \ac{mae}, its higher \ac{rss} and considerably higher \ac{kl} indicate a posterior that deviates further from the prior without any gain in accuracy, suggesting less well-calibrated uncertainty modeling. LCModel shows a substantially higher \ac{mae} relative to the data-driven models; note, however, that LCModel is unaware of the simulation ranges and uses its own baseline model, so part of this gap reflects model mismatch rather than the estimator itself.

Fig.~\ref{fig:calibration} shows the calibration curves of the three models, comparing nominal coverage (1\% to 99\%) to empirical coverage, i.e., the proportion of true values that fall within the predicted intervals. %The ideal calibration curve follows a 1:1 line (shown as the dashed black line), and the curves of the models indicate how close the predicted intervals are to this ideal.
For LCModel, a Gaussian with the \ac{crb} as standard deviation is used as the predictive distribution.
The \ac{snf} is well calibrated for all metabolites, with coverage close to the ideal line over the entire range. The \ac{vae}, despite matching the \ac{snf} in point accuracy, is markedly over-confident for a subset of strongly overlapping or low-concentration metabolites (e.g., \ac{cr}, \ac{pcr}, \ac{gpc}, and \ac{pch}), where a nominal 90\% interval covers the true value in less than half of the cases; a diagonal Gaussian cannot represent the anti-correlated posteriors of Fig.~\ref{fig:fits_n_latent} and therefore underestimates the marginal spread. LCModel, based on the \ac{crb}, is over-confident for every metabolite, in line with the \ac{crb} providing only a theoretical lower bound on the error rather than a calibrated uncertainty estimate.

%%%%%%%%%%%%%%%%%%
%%% Discussion %%%
%%%%%%%%%%%%%%%%%%
\vspace{-1mm}
\section{Conclusion} \label{sec:conclusion}
\vspace{-2mm}
% Previous version:
% This study shows that flow-based models can learn meaningful posterior distributions for metabolite quantification from \ac{mrs} data. Beyond achieving good predictive accuracy and uncertainty calibration, the learned posteriors reflect properties of the spectral information, including correlations, ambiguities, and dependencies between metabolites. Ambiguities of the quantification task can thus be read directly from the model output. Overall, variational Bayesian inference with flow-based posteriors provides a viable path toward more informative uncertainty estimates; validation on in vivo data is the subject of future work.
We presented a physics-informed variational inference framework for \ac{mrs} quantification, in which a \ac{snf} approximates the posterior over the parameters of an explicit \ac{mrs} signal model. Replacing the learned decoder by the physics-based forward model ties the latent space to physically meaningful quantities, allows prior knowledge on the signal to be built in directly, and, since no labels are required, lets the model be trained and applied directly on in vivo spectra. On simulated 7T data, the flow-based posteriors match a \ac{vae} in accuracy but are markedly better calibrated and expose the structure of the inverse problem, including anti-correlated posteriors that the \ac{crb} cannot represent.

%%%%%%%%%%%%%%%%%%
%%% References %%%
%%%%%%%%%%%%%%%%%%
\vspace{-1mm}
\bibliographystyle{styles/IEEEbib}
\bibliography{refs/lr_refs, refs/my_refs, refs/refs}

@book{de_graaf_vivo_2019,
  title = {In Vivo {{NMR}} Spectroscopy: Principles and Techniques},
  shorttitle = {In Vivo {{NMR}} Spectroscopy},
  author = {De Graaf, Robin A.},
  year = {2019},
  edition = {3rd ed},
  publisher = {{John Wiley \& Sons, Inc}},
  address = {{Hoboken, NJ}},
  isbn = {978-1-119-38251-5 978-1-119-38246-1 978-1-119-38257-7},
  langid = {english},
  lccn = {616.075 48}
}

@article{faghihi_magnetic_2017,
  title = {Magnetic {{Resonance Spectroscopy}} and Its {{Clinical Applications}}: {{A Review}}},
  shorttitle = {Magnetic {{Resonance Spectroscopy}} and Its {{Clinical Applications}}},
  author = {Faghihi, Reza and {Zeinali-Rafsanjani}, Banafsheh and {Mosleh-Shirazi}, Mohammad-Amin and {Saeedi-Moghadam}, Mahdi and Lotfi, Mehrzad and Jalli, Reza and Iravani, Vida},
  year = {2017},
  month = sep,
  journal = {Journal of Medical Imaging and Radiation Sciences},
  volume = {48},
  number = {3},
  pages = {233--253},
  issn = {1939-8654},
  doi = {10.1016/j.jmir.2017.06.004},
  urldate = {2022-04-04},
  langid = {english}
}

@article{near_preprocessing_2021,
  title = {Preprocessing, Analysis and Quantification in Single-Voxel Magnetic Resonance Spectroscopy: Experts' Consensus Recommendations},
  shorttitle = {Preprocessing, Analysis and Quantification in Single-Voxel Magnetic Resonance Spectroscopy},
  author = {Near, Jamie and Harris, Ashley D. and Juchem, Christoph and Kreis, Roland and Marja{\'n}ska, Ma{\l}gorzata and {\"O}z, G{\"u}lin and Slotboom, Johannes and Wilson, Martin and Gasparovic, Charles},
  year = {2021},
  journal = {NMR in Biomedicine},
  volume = {34},
  number = {5},
  pages = {e4257},
  issn = {1099-1492},
  doi = {10.1002/nbm.4257},
  urldate = {2022-02-08},
  langid = {english}
}

@article{provencher_estimation_1993,
  title = {Estimation of Metabolite Concentrations from Localizedin Vivo Proton {{NMR}} Spectra},
  author = {Provencher, Stephen W.},
  year = {1993},
  month = dec,
  journal = {Magnetic Resonance in Medicine},
  volume = {30},
  number = {6},
  pages = {672--679},
  issn = {07403194, 15222594},
  doi = {10.1002/mrm.1910300604},
  urldate = {2023-01-31},
  langid = {english}
}

@inproceedings{Higgins2017betaVAE,
  author    = {Irina Higgins and Loic Matthey and Arka Pal and Christopher Burgess and Xavier Glorot and Matthew Botvinick and Shakir Mohamed and Alexander Lerchner},
  title     = {beta-{VAE}: Learning Basic Visual Concepts with a Constrained Variational Framework},
  booktitle = {International Conference on Learning Representations (ICLR)},
  year      = {2017}
}

@inbook{Hurd2009ArtifactsAPI, 
    place={Cambridge}, 
    title={Artifacts and pitfalls in MR spectroscopy}, 
    booktitle={Clinical MR Neuroimaging: Physiological and Functional Techniques}, 
    publisher={Cambridge University Press}, 
    author={Hurd, Ralph E.}, 
    editor={Gillard, Jonathan H. and Waldman, Adam D. and Barker, Peter B.Editors}, 
    year={2009}, 
    pages={30–43}
}

@Inbook{Horska2023MRSClincalA,
author="Horsk{\'a}, Alena
and Berrington, Adam
and Barker, Peter B.
and Tk{\'a}{\v{c}}, Ivan",
editor="Faro, Scott H.
and Mohamed, Feroze B.",
title="Magnetic Resonance Spectroscopy: Clinical Applications",
bookTitle="Functional Neuroradiology: Principles and Clinical Applications",
year="2023",
publisher="Springer International Publishing",
address="Cham",
pages="241--292",
isbn="978-3-031-10909-6",
doi="10.1007/978-3-031-10909-6_10",
url="https://doi.org/10.1007/978-3-031-10909-6_10"
}

@article{Kreis2004IssuesOS,
  title={Issues of spectral quality in clinical 1H‐magnetic resonance spectroscopy and a gallery of artifacts},
  author={Roland Kreis},
  journal={NMR in Biomedicine},
  year={2004},
  volume={17},
  url={https://api.semanticscholar.org/CorpusID:28127837}
}

@article{Maudsley2020AdvancedMR,
  title={Advanced magnetic resonance spectroscopic neuroimaging: Experts' consensus recommendations},
  author={Andrew A. Maudsley and Ovidiu C. Andronesi and Peter B. Barker and Alberto Bizzi and Wolfgang Bogner and Anke Henning and Sarah J. Nelson and Stefan Posse and Dikoma C. Shungu and Brian J. Soher},
  journal={NMR in Biomedicine},
  year={2020},
  volume={34},
  url={https://api.semanticscholar.org/CorpusID:217593781}
}

@article{Zhang2017FastSpectroscopy,
    title = {{Fast computation of full density matrix of multispin systems for spatially localized in vivo magnetic resonance spectroscopy}},
    year = {2017},
    journal = {Medical Physics},
    author = {Zhang, Yan and An, Li and Shen, Jun},
    pages = {4169--4178},
    volume = {44}
}

@article{Landheer2021AreCRLBs,
author = {Landheer, Karl and Juchem, Christoph},
title = {Are Cramér-Rao lower bounds an accurate estimate for standard deviations in in vivo magnetic resonance spectroscopy?},
journal = {NMR in Biomedicine},
volume = {34},
number = {7},
pages = {e4521},
doi = {https://doi.org/10.1002/nbm.4521},
url = {https://analyticalsciencejournals.onlinelibrary.wiley.com/doi/abs/10.1002/nbm.4521},
eprint = {https://analyticalsciencejournals.onlinelibrary.wiley.com/doi/pdf/10.1002/nbm.4521},
year = {2021}
}

@article{Condon2011MagneticRI,
  title={Magnetic resonance imaging and spectroscopy: how useful is it for prediction and prognosis?},
  author={Barrie Condon},
  journal={The EPMA Journal},
  year={2011},
  volume={2},
  pages={403 - 410},
  url={https://api.semanticscholar.org/CorpusID:13221168}
}

@InProceedings{Rezende2015normFlows,
  title = 	 {Variational Inference with Normalizing Flows},
  author = 	 {Rezende, Danilo and Mohamed, Shakir},
  booktitle = 	 {Proceedings of the 32nd International Conference on Machine Learning},
  pages = 	 {1530--1538},
  year = 	 {2015},
  volume = 	 {37},
  url = 	 {https://proceedings.mlr.press/v37/rezende15.html}
}

@inproceedings{vandenBerg2018SNFs,
title = "Sylvester normalizing flows for variational inference",
author = "{Van Den Berg}, Rianne and Leonard Hasenclever and Tomczak, {Jakub M.} and Max Welling",
year = "2018",
month = jan,
day = "1",
language = "English",
pages = "393--402",
booktitle = "34th Conference on Uncertainty in Artificial Intelligence 2018, UAI 2018",
}

@article{Kobyzev2020NormalizingFA,
  title={Normalizing Flows: An Introduction and Review of Current Methods},
  author={Ivan Kobyzev and Simon Prince and Marcus A. Brubaker},
  journal={IEEE Transactions on Pattern Analysis and Machine Intelligence},
  year={2020},
  volume={43},
  pages={3964-3979},
  url={https://api.semanticscholar.org/CorpusID:208910764}
}

@inproceedings{Kingma2014VAE,
  author = {Kingma, Diederik P. and Welling, Max},
  booktitle = {2nd International Conference on Learning Representations, {ICLR} 2014},
  eprint = {http://arxiv.org/abs/1312.6114v10},
  eprintclass = {stat.ML},
  eprinttype = {arXiv},
  title = {{Auto-Encoding Variational Bayes}},
  year = 2014
}

\end{document}